\documentclass[10pt,twocolumn,letterpaper]{article}

\usepackage[pagenumbers]{cvpr}
\usepackage{hyperref}
\hypersetup{
    colorlinks=true,
    citecolor=MidnightBlue,
    linkcolor=OrangeRed,
    urlcolor=OrangeRed
}
\usepackage{multirow}
\usepackage{float}

\title{Driv3R: Learning Dense 4D Reconstruction for Autonomous Driving}

\author{Xin Fei$^{1,2,}$\footnotemark[1] \quad 
Wenzhao Zheng$^{1,2,}$\footnotemark[2] \quad 
Yueqi Duan$^1$ \quad 
Wei Zhan$^2$ \quad \\
{Masayoshi Tomizuka}$^2$ \quad 
{Kurt Keutzer}$^2$ \quad
{Jiwen Lu}$^1$ \\
$^1$Tsinghua University \quad 
$^2$University of California, Berkeley \\
\texttt{xinfei\_21@hotmail.com; wenzhao.zheng@outlook.com} \\
Project Page: \url{https://wzzheng.net/Driv3R} \\
Large Driving Models: \url{https:/github.com/wzzheng/LDM}
}

\begin{document}

\twocolumn[{%
\renewcommand\twocolumn[1][]{#1}
\vspace{-7mm}
\maketitle
\vspace{-11mm}
\begin{center}
   \centering
   \includegraphics[width=\linewidth]{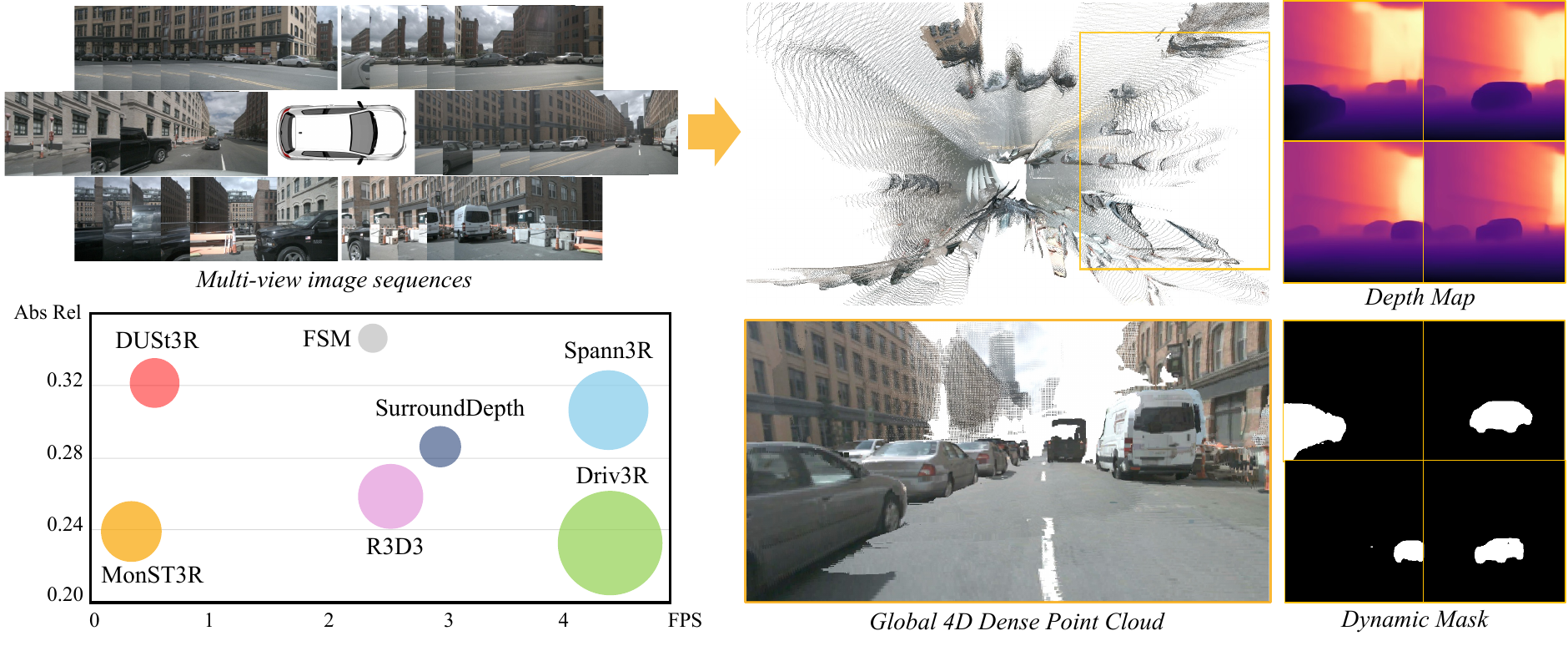}
   \vspace{-7mm}
   \captionof{figure}{
   Our Driv3R predicts dense 4D dynamic point clouds in the global world coordinate system in a streaming manner from multi-view images. 
   It outperforms existing methods in reconstructing dynamic autonomous driving scenes and achieves a $15 \times$ faster inference speed compared to approaches that require global alignment optimization~\cite{zhang2024monst3r, dust3r_cvpr24}.
}
\label{teaser}
\end{center}%
}]

\renewcommand{\thefootnote}{\fnsymbol{footnote}}
\footnotetext[1]{Work done while visiting UC Berkeley. $\dagger$Corresponding author.}
\renewcommand{\thefootnote}{\arabic{footnote}}

\begin{abstract}
    Realtime 4D reconstruction for dynamic scenes remains a crucial challenge for autonomous driving perception. 
    Most existing methods rely on depth estimation through self-supervision or multi-modality sensor fusion. 
    In this paper, we propose \textbf{Driv3R}, a DUSt3R-based framework that directly regresses per-frame point maps from multi-view image sequences. 
    To achieve streaming dense reconstruction, we maintain a memory pool to reason both spatial relationships across sensors and dynamic temporal contexts to enhance multi-view 3D consistency and temporal integration. 
    Furthermore, we employ a 4D flow predictor to identify moving objects within the scene to direct our network focus more on reconstructing these dynamic regions. 
    Finally, we align all per-frame pointmaps consistently to the world coordinate system in an optimization-free manner. 
    We conduct extensive experiments on the large-scale nuScenes dataset to evaluate the effectiveness of our method.
     Driv3R outperforms previous frameworks in 4D dynamic scene reconstruction, achieving $15 \times$ faster inference speed compared to methods requiring global alignment. 
     Code: \url{https://github.com/Barrybarry-Smith/Driv3R}.
\end{abstract}    
\vspace{-0.3cm}
\section{Introduction}

Real-time and accurate dense reconstruction of dynamic scenes is a challenging task for the perception of autonomous driving and robotics. Compared to data fusion from multi-modality sensors, such as cameras, LiDAR, and radar, relying solely on multi-view cameras provides a more computationally efficient and low-cost solution. However, achieving accurate depth estimation without 3D ground truth supervision and precise representations of dynamic objects introduces significant challenges to this task. 

To address these challenges, several efficient 3D representations have been proposed to enable scene reconstruction from multi-view cameras and perform downstream tasks, such as novel view synthesis, depth estimation, and pose prediction. Mildenhall et al.~\cite{mildenhall2020nerf} encode multi-view image inputs into implicit neural radiance field (NeRF) for 3D representations, followed by works improving the efficiency, performance, and generalizability of NeRF model~\cite{Hu_2022_CVPR, xu2024murf, yu2021pixelnerf, chen2021mvsnerf}. More recently, explicit 3D Gaussian representation has been demonstrated to achieve better performance and significantly improved efficiency with the rasterization-based renderer~\cite{kerbl3Dgaussians, charatan23pixelsplat, chen2024mvsplat, wewer24latentsplat, fei2024pixelgaussiangeneralizable3dgaussian}. Meanwhile, to adapt the origin NeRF and 3D Gaussian representation to model the complex dynamic objects and scenes more accurately, several works have explored incorporating strategies such as object tracking, optical flow, and motion encoding to further extend the 3D representations to support dynamic 4D reconstruction~\cite{Gao-ICCV-DynNeRF, liu2024modgsdynamicgaussiansplatting, lei2024moscadynamicgaussianfusion}.

Despite the advances in efficient 3D representations, precise and dense point cloud remains to be highly significant for reconstruction, particularly in the context of autonomous driving. In this direction, DUSt3R~\cite{dust3r_cvpr24} pioneers learning strong 3D priors solely from input image pairs, which directly regress pixel-aligned point cloud representations and confidence maps. Furthermore, MonST3R~\cite{zhang2024monst3r} extends the DUSt3R representations to model dynamic scenes by strategically fine-tuning on appropriate datasets. However, MonST3R still relies on the computationally expensive global alignment process used in DUSt3R and struggles to efficiently model dynamic large-scale scenes in autonomous driving. To eliminate the need for such alignment proposed by DUSt3R, Spann3R~\cite{wang20243d} introduces a spatial memory pool to update features encoded by ViT~\cite{dosovitskiy2020image}, enabling an incremental 3D reconstruction process within a consistent coordinate system. However, such spatial memory can only reason spatial relationships in static scenes and is not able to effectively handle temporal information fusion, thus limiting its ability to reconstruct dynamic and large-scale scenes in the context of autonomous driving.

To address this, we propose a \textbf{Driv3R} model to reconstruct large-scale dynamic autonomous driving scenes in the global world coordinate system without global alignment optimization. Specifically, we maintain a memory pool to reason both temporal relationships and spatial contexts within the multi-view sequences. Additionally, to accurately capture dynamic objects from input images, we adopt the lightweight RAFT~\cite{teed2020raftrecurrentallpairsfield} model followed by segmentation refinement in the 4D flow predictor. Having generated masks for moving objects with the flow predictor, we adopt the point cloud predictions from the pretrained R3D3~\cite{r3d3} model as supervision to direct our model to focus more on these dynamic regions for accurate 4D reconstruction. Furthermore, by ensuring multi-view 3D consistency through information interactions between different viewpoints in the temporal-spatial memory pool, we can align the per-frame point maps to the world coordinate system in an optimization-free manner, ultimately reconstructing the complete large-scale 4D dynamic autonomous driving scenes. We conduct extensive experiments on the large-scale nuScenes~\cite{nuscenes2019} dataset for depth estimation and scene reconstruction, where Driv3R achieves results comparable to state-of-the-art multi-view depth estimation frameworks. In addition, for 4D reconstruction on dynamic large-scale scenes, our method outperforms all existing methods with a $15 \times$ faster inference speed compared to methods that rely on global alignment optimization.
\section{Related Work}

\textbf{Depth Estimation For Autonomous Driving.} Due to the absence of dense ground-truth depth in large-scale autonomous driving datasets, previous research either adopts a self-supervised approach for depth estimation~\cite{zhou2017unsupervisedlearningdepthegomotion, monodepth2, bian2019unsupervisedscaleconsistentdepthegomotion, zhao2020towards, yin2018geonet, zhou2019movingindoorunsupervisedvideo, ranjan2019competitivecollaborationjointunsupervised, r3d3, wei2022surround} or incorporates additional supervisory signals, such as LiDAR~\cite{feiWS19, kuznietsov2017semisuperviseddeeplearningmonocular}, optical flow~\cite{zou2018dfnet, yin2018geonet} and object motion~\cite{casser2018depthpredictionsensorsleveraging, ranjan2019competitivecollaborationjointunsupervised}, to enhance prediction accuracy. Among these methods, R3D3~\cite{r3d3} leverages both temporal and spatial information from multi-view cameras by iterating between geometric estimations and further refines the monocular depth, enabling accurate and efficient dense depth prediction in dynamic scenes. We thus adopt the depth predictions from the pretrained R3D3 model as supervision for moving objects in our Driv3R model.

\noindent\textbf{Static 3D Reconstruction.} Static 3D reconstruction for objects and scenes has seen considerable advancement with the rise of learning-based approaches. These methods aim to learn meshes~\cite{gkioxari2020meshrcnn, wang2018pixel2meshgenerating3dmesh, pontes2017image2mesh, kato2018renderer}, point clouds~\cite{guo2020deeplearning3dpoint, lin2017learningefficientpointcloud, dust3r_cvpr24, wang20243d}, voxels~\cite{sitzmann2019deepvoxels, choy20163d, drcTulsiani17}, implicit neural fields~\cite{mildenhall2020nerf, yu2021pixelnerf, chen2021mvsnerf, Hu_2022_CVPR, xu2024murf} or explicit representations~\cite{kerbl3Dgaussians, charatan23pixelsplat, chen2024mvsplat, szymanowicz24splatter, fei2024pixelgaussiangeneralizable3dgaussian} from the training data. DUSt3R~\cite{dust3r_cvpr24} takes a pioneering step to directly regress point maps from arbitrary input image pairs leveraging strong 3D priors learned from the large-scale training data. However, DUSt3R requires computationally expensive global alignment to optimize pointmaps and camera poses into a consistent coordinate system. To address this, Spann3R~\cite{wang20243d} maintains a spatial memory pool that enables incremental 3D reconstruction in a consistent coordinate system from input sequences, eliminating the need for alignment optimization. However, these methods perform poorly to reconstruct dynamic scenes, which are crucial for autonomous driving perception.

\noindent\textbf{Optical Flow.} Identifying dynamic objects is crucial for the accurate reconstruction of 4D scenes. Optical flow, which estimates per-pixel motion across image sequences, plays a key role in detecting moving objects in 2D images. Previously, the estimation of optical flow is treated as an energy minimization process~\cite{HORN1981185} or discrete optimization problem~\cite{chen2016flowopticalflowestimation, xu2017accurateopticalflowdirect}, while more recent works tend to adopt end-to-end differentiable neural networks for improved efficiency and accuracy~\cite{Bar-Haim_2020_CVPR, hofinger2020improvingopticalflowpyramid, hui18liteflownet, Hur:2019:IRR, teed2020raftrecurrentallpairsfield}. The lightweight RAFT~\cite{teed2020raftrecurrentallpairsfield} model constructs multi-scale 4D correlation volumes and updates the flow field using a recurrent network, achieving outstanding performance in optical flow prediction with high inference efficiency. Therefore, we adopt RAFT as the core component in our 4d flow predictor.

\noindent\textbf{Dynamic 4D Reconstruction.} Since the introduction of NeRF~\cite{mildenhall2020nerf}, subsequent researchers have extended its implicit neural field representations to enable novel view synthesis in 4D dynamic scenes~\cite{Gao-ICCV-DynNeRF, tonderski2024neuradneuralrenderingautonomous, tonderski2024neuradneuralrenderingautonomous}. More recently, 3D Gaussian has been explored as an efficient explicit representation for scene reconstruction~\cite{kerbl3Dgaussians}. Therefore, several works have focused on leveraging Gaussian representations to encode the motions and deformations in dynamic scenes for real-time rendering~\cite{Wu_2024_CVPR} and novel view synthesis based on monocular videos~\cite{lei2024moscadynamicgaussianfusion, liu2024modgsdynamicgaussiansplatting}. Additionally, GFlow~\cite{wang2024gflowrecovering4dworld} and DreamScene4D~\cite{dreamscene4d} manage to reconstruct dynamic scenes from monocular videos without camera parameters, enhancing the scene recovery and object tracking for in-the-wild scenarios. Moreover, MonST3R~\cite{zhang2024monst3r} proposes to directly estimate per-timestamp geometry for each frame, successfully adapting DUSt3R~\cite{dust3r_cvpr24} representations to reconstruct dynamic scenes. However, MonST3R still depends on the computationally expensive global alignment optimization proposed by DUSt3R and struggles to efficiently represent the entire scene within a consistent coordinate system. In this work, we take a step further to model the 4D dynamic scenes without any optimization, enabling real-time scene reconstruction for autonomous driving.

\section{Proposed Approach}

\subsection{Problem Formulation}
Given multi-view images $\mathcal{I}_t = \{I_{t, c}\}_{c=1}^{C}$ from RGB cameras with corresponding camera poses $\{\mathbf{T}_{t, c}, \mathbf{K}_{t, c} \}_{c=1}^{C}$ for each timestamp $t \in \mathcal{T}$, our Driv3R aims to learn 4D dense pixel-wise point cloud representations $\{P_t \mid t \in \mathcal{T} \}$ in the global world coordinate system.

Our Driv3R is composed of three stages and allows for end-to-end training. First, we construct a memory pool for spatial and temporal information interaction inspired by Spann3R~\cite{wang20243d}. Then, we introduce a 4d flow predictor to identify dynamic objects within the scene, guiding the network to focus more on reconstructing these regions during training. Finally, we employ a multi-view aligner to register all point maps into the consistent world coordinate system in an optimization-free manner. Leveraging the depth inference from the pretrained R3D3~\cite{r3d3} model as supervision, Driv3R enables consistent 4D dense point cloud reconstruction and precise modeling of moving objects within the dynamic scenes.

\subsection{Temporal-Spatial Memory Pool}
\label{sec:3.1}
Spann3R~\cite{wang20243d} takes multi-view images as input and manages an external spatial memory to predict per-image pointmaps in a consistent coordinate system. Inspired by this, we maintain a temporal-spatial memory pool to reason both temporal relationships and spatial contexts within the multi-view input sequences. To elaborate, given frames from various sensors and timestamps in the input sequence, denoted as $I_{t, c}$ and $I_{t', c'}$, a ViT~\cite{dosovitskiy2020image} first encodes both images into feature maps $f_{t, c}, f_{t', c'}$. Then, we update $f_{t, c}$ by extracting memories from the temporal-spatial memory pools as described in Equation~\ref{eq:temporal_spatial_memory}, which allows the encoded feature $f_{t, c}$ to fully interact with information obtained from previous timestamps and viewpoints.

\begin{equation}
\label{eq:temporal_spatial_memory}
    f_{t, c}^{*} = \textit{softmax}(\frac{q_{t, c} \ K^{T}}{\sqrt{s}} V) + q_{t, c},
\end{equation}
where $q_{t, c}, K, V$ represent the query for the current frame, and the key and value from memory pools, respectively. Subsequently, we decode the feature pair $(f_{t, c}^{*}, f_{t', c'})$ via two interconnected decoders. The feature decoded by the target decoder is used to generate the query for the next step, while the feature decoded by the reference decoder is applied to memory encoding and regression of per-frame point maps and confidence maps (as shown in Equation~\ref{eq:decoder}).

\begin{equation}
\label{eq:decoder}
\begin{gathered}
    f_{t, c}^{tar}, f_{t, c}^{ref} = \textit{Decoder}(f_{t, c}^{*}, f_{t', c'}), \\
    P_{t, c}, C_{t, c} = \textit{MLP}(f_{t, c}^{ref}), \\
    K_{t, c}, V_{t, c} = \textit{Encoder}(f_{t, c}^{ref}, f_{t, c}, P_{t, c}).
\end{gathered}
\end{equation}

\begin{figure}[t]
    \centering
    \includegraphics[width=\linewidth]{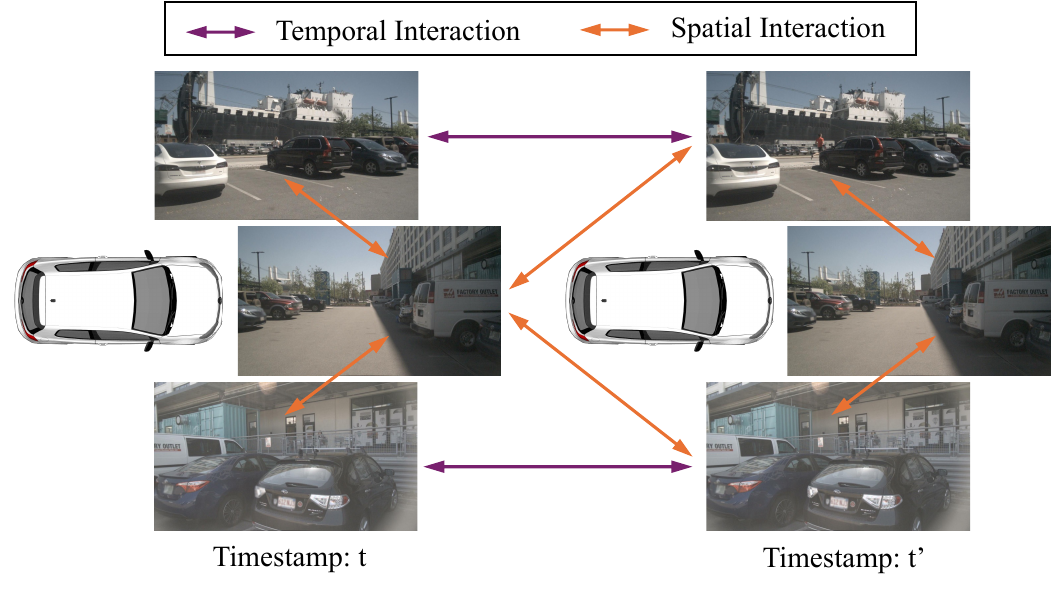}
	\vspace{-7mm}
    \caption{\textbf{Temporal and spatial interactions within the memory pool.} By maintaining a sensor-aware memory pool where key-value pairs are stored in the order of timestamps, we clearly identify both spatial and temporal relationships and perform a more efficient feature update process.}
	\vspace{-5mm}
    \label{fig:spatial_memory_pool}
\end{figure}

\begin{figure*}[t]
    \centering
    \includegraphics[width=1\linewidth]{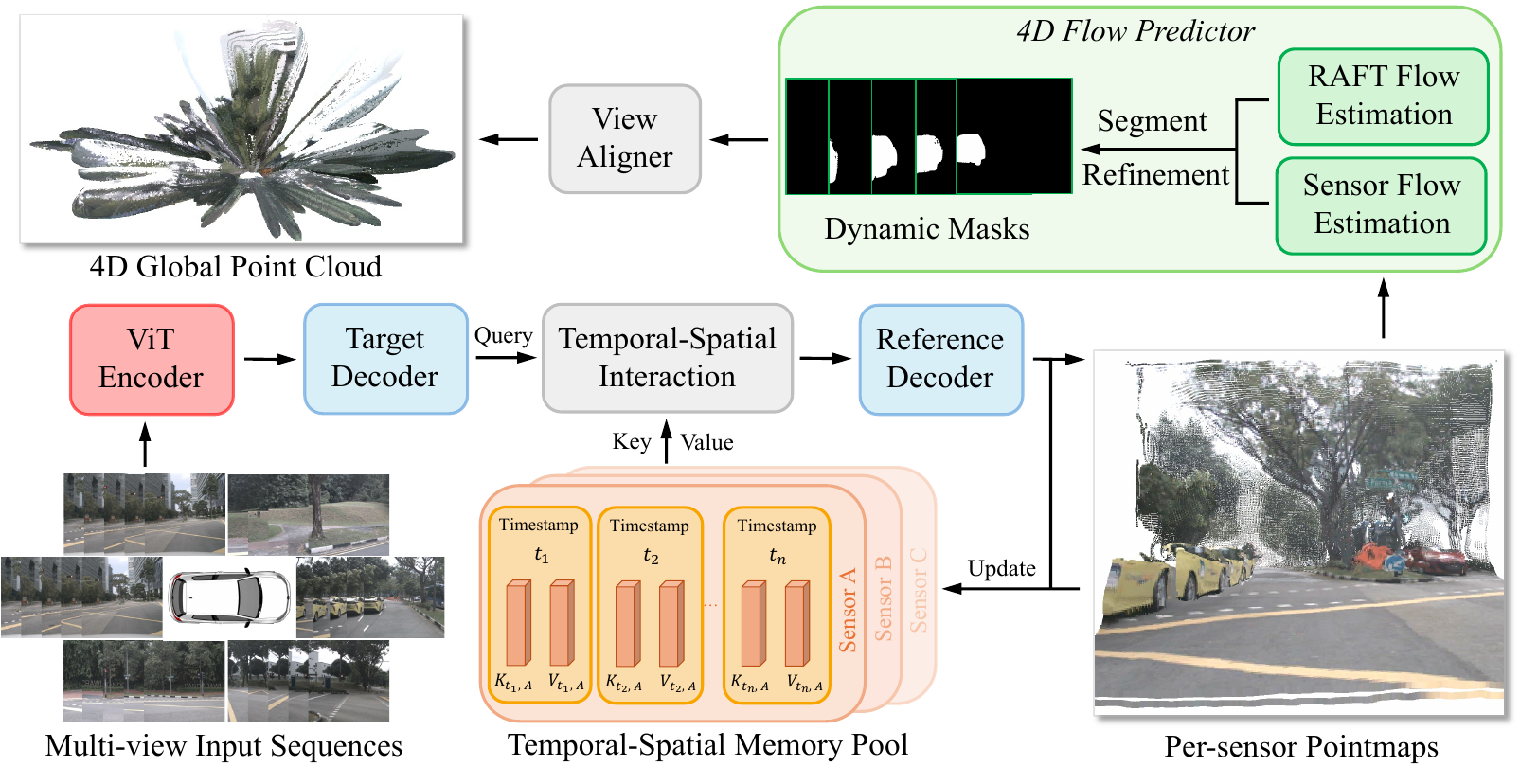}
	\vspace{-7mm}
    \caption{\textbf{Overview of Driv3R.} Given multi-view input sequences, we construct a sensor-wise memory pool for temporal and spatial information interactions. After obtaining per-frame point maps, the 4D flow predictor identifies the dynamic objects within the scene. Finally, we adopt an optimization-free multi-view alignment strategy to predict the 4D global point cloud in the world coordinate system.}
	\vspace{-5mm}
    \label{fig:pipeline}
\end{figure*}

Compared to Spann3R~\cite{wang20243d}, our architecture efficiently handles both spatial and temporal information storage, management, and interaction. Specifically, we maintain a separate memory pool for each sensor with each key-value pair labeled by its corresponding timestamp. During pool updates, new keys are added to the relevant sensor pool based on cosine similarities with the existing memory keys, allowing us to identify which key-value pairs have the closest spatial and temporal relationships with the current frame when using the memory pool to update the feature generated from ViT (as shown in Figure~\ref{fig:spatial_memory_pool}). By performing cross-attention only on relevant key-value pairs, we effectively reduce unnecessary computational overheads and minimize interference from irrelevant frames.

\subsection{4D Flow Predictor}
\label{sec:3.2}

To enhance the ability of our Driv3R model to reconstruct dynamic objects in the input scene, we design a 4D flow predictor based on RAFT~\cite{teed2020raftrecurrentallpairsfield} model. Given image sequence from a single sensor $\{I_{1}, I_{2}, ..., I_{T}\}$ and corresponding pointmaps $\{P_{1}, P_{2}, ..., P_{T}\}$ obtained in Section~\ref{sec:3.2}, we first generate a set of pairs $\{(I_{i_1}, I_{i_2})\}_{i=1}^{M}$ that include frames which are temporally adjacent. Then, we use the pretrained RAFT model to predict flow maps $\{(F_{12}^{i}, F_{21}^{i})\}_{i=1}^{M}$ for each image pair. To further capture the 4D motions of objects, we apply cross-projection on the pointmaps to obtain the flow map induced by the motion of the sensor (as shown in Equation~\ref{eq:project_to_get_flow}). Thus, we derive the coarse dynamic mask $\{\Omega_1^{'}, \Omega_2^{'}, ..., \Omega_T^{'}\}$ for each frame by simply averaging the corresponding masks across all frame pairs.

Next, we incorporate the pretrained SAM2~\cite{ravi2024sam2} model for segmentation to further refine the coarse dynamic masks. To illustrate, for each mask at frame $t$, we first binarize $\Omega_{t}^{'}$ and feed each connected mask region along with the original image into SAM2, using the segmentation output to augment the binary coarse mask. This helps to fill in the missing parts of the initial mask, thereby ensuring comprehensive coverage of the dynamic objects. Finally, we obtain the refined masks of dynamic objects $\{\Omega_1, \Omega_2, ..., \Omega_T\}$. The overall process of 4D flow predictor can be formulated as:

\begin{equation}
\label{eq:project_to_get_flow}
\begin{gathered}
    E_{12}^{i} = K_{i_2}^{(e)}{T_{i_2}^{(e)}}^{-1}P_{i_1}^{(e)} - K_{i_1}^{(e)}{T_{i_1}^{(e)}}^{-1}P_{i_1}^{(e)}, \\
    K_{i_j}^{(e)}, T_{i_j}^{(e)} = \textit{PoseEstimate}(P_{i_j}), \ j = 1, 2.
\end{gathered}
\end{equation}

\vspace{-0.3cm}
\begin{equation}
\label{eq:4d_flow_predictor}
\begin{gathered}
\Omega_{t}^{'} = \frac{1}{N} (\sum_{i_1 = t} {||F_{12}^{i} - E_{12}^{i}||} + \sum_{i_2 = t} {||F_{21}^{i} - E_{21}^{i}||}), \\
\Omega_{t} = \mathcal{S}(\mathcal{B}(\Omega_{t}^{'}), I_t) + \mathcal{B}(\Omega_{t}^{'}),
\end{gathered}
\end{equation}
where \(\textit{PoseEstimate}(\cdot)\) denotes the estimation of camera extrinsics and intrinsics from the given point map as described in DUSt3R~\cite{dust3r_cvpr24}. $\mathcal{S}(\cdot)$ and $\mathcal{B}(\cdot)$ stand for SAM2 augmentation and binary operation, respectively.

\subsection{Multi-view Aligner}
\label{sec:3.3}

After regressing per-frame point cloud predictions from multi-view input sequences, pointmaps from multiple sensors, denoted as $\{P_{t, c}, t \in \mathcal{T}, c = 1, 2, ..., C\}$, are initially represented within their respective coordinate systems. Therefore, we employ a multi-view aligner to align these point maps into the global world coordinate system in an optimization-free manner. Specifically, we first obtain the camera parameters through pose estimation for each frame within the respective coordinate system of its input sequence, which allows us to project the predicted point maps into per-frame depth maps. Therefore, each depth map can be unprojected to points in the global world coordinate system using the ground truth camera parameters (as illustrated in Equation~\ref{eq:multi_view_aligner}). Since all the per-frame point maps are decoded from features that fully capture both temporal and spatial information within the memory pool, such simple pose transformation can ensure point cloud consistency both temporally and spatially. Finally, we obtain per-frame dense point maps $\{P_t \mid t \in \mathcal{T} \}$ from multi-sensor inputs within the real-world coordinate system without requiring any additional alignment:
\begin{equation}
\label{eq:multi_view_aligner}
\begin{gathered}
    K_{t, c}^{(e)}, T_{t, c}^{(e)} = \textit{PoseEstimate}(P_{t, c}), \\
    P_{t, c}^{(w)} = \mathbf{T_{t, c}} {\mathbf{K_{t, c}}}^{-1} K_{t, c}^{(e)} {T_{t, c}^{(e)}}^{-1} P_{t, c}.
\end{gathered}
\end{equation}

\subsection{Training of Driv3R}
\label{sec:3.4}

\textbf{Loss.} Due to the sparsity of point clouds obtained from LiDAR in autonomous driving, we use the dense depth estimation results from the pretrained R3D3~\cite{r3d3} model as supervision. After the warm-up training steps, we supervise only the points aligned with dynamic objects, which are identified by the 4d flow predictor in Section~\ref{sec:3.2}, to calculate the confidence-aware loss (as shown in Equation~\ref{eq:training_loss}). Following DUSt3R~\cite{dust3r_cvpr24}, both the predicted and ``ground truth" point maps from R3D3 prediction are normalized by average distances. Furthermore, we also add a scale loss to encourage the scale of predicted point maps to be smaller than the predicted ones from the pretrained R3D3 model.

\begin{equation}
\label{eq:training_loss}
\mathcal{L} = \mathcal{L}_{conf}^{dynamic} + \mathcal{L}_{scale}^{dynamic}.
\end{equation}

\noindent\textbf{Two-stage Training.} Our Driv3R model is trained in two stages to address the limitations of training memory. In the first stage, the input sequences consist of images from a single sensor, which means the memory pool is used solely for temporal information interactions. In the second stage, the input sequences consist of images from different sensors and timestamps with overlapping fields of view, which allows the memory pool to further reason about spatial relations and ensures the output pointmaps are spatially consistent. Finally, we adopt the optimization-free multi-view aligner to formulate the complete 4D dense point cloud predictions in the global world coordinate system.
\section{Experiments}

\subsection{Datasets} 

The nuScenes~\cite{nuscenes2019} dataset consists of 1,000 diverse driving scenes, each lasting approximately 20 seconds with keyframes annotated at a frequency of 2Hz. 
The 1,000 scenes in nuScenes are officially split into 700 for training, 150 for validation, and 150 for testing.
 Each keyframe contains RGB cameras from six surrounding cameras and a sparse point cloud collected from lidar. We divide each sample into sequences in the temporal order and group consecutive sets of 5 frames as inputs to Driv3R. 
We only use multi-view camera data during training and inference and adopt the pretrained R3D3~\cite{r3d3} model to generate noisy depth supervision.

\subsection{Implementation Details} 

Given that the initial resolution of images in nuScenes~\cite{nuscenes2019} is 1600x900, we split each camera into two virtual cameras with a resolution of 224x224, so that we can leverage the pretrained memory encoder in Spann3R~\cite{wang20243d} and ensure the multi-view inputs are able to cover the entire panoramic fields of view. Our Driv3R model is trained on 8 NVIDIA A6000 GPUs with a batch size of 4. As described in Section~\ref{sec:3.4}, our model is trained in two stages to address memory limitations. We first train Driv3R for 30 epochs using temporal sequences. Then, we fine-tune the model for 20 additional epochs with a small learning rate, allowing the memory pool to further reason spatial relationships in sequences composed of multi-view images from different timestamps. For fair comparisons, we fine-tune all baseline models not trained on nuScenes~\cite{nuscenes2019} for the same number of epochs using predicted points from R3D3~\cite{r3d3} as supervision. We provide more details in Sec~\ref{app:training}.

\subsection{Evaluation Metrics}

Following previous work in dense 3D reconstruction~\cite{wang20243d, Azinovic_2022_CVPR}, we use \textbf{accuracy}, \textbf{completion}, and \textbf{normal consistency} to evaluate the quality of reconstruction. Additionally, we also calculate the standard metrics to assess the quality of depth estimation, including \textit{Abs Rel}, \textit{Sq Rel}, \textit{RMSE}, and the percentage of prediction with $\delta < 1.25$, $\delta < 1.25^2$ and $\delta < 1.25^3$. The predicted point maps and ground-truth LiDAR point clouds are transformed into a consistent global coordinate system for the assessment of reconstruction quality, and into respective sensor coordinate systems for the evaluation on depth predictions.

\begin{table*}[t]
    \centering
    \caption{\textbf{Depth evaluation on the nuScenes~\cite{nuscenes2019} validation set.} For FSM~\cite{guizilini2021surroundmonodepthmultiplecameras}, R3D3~\cite{r3d3} and SurroundDepth~\cite{wei2022surround}, all images are input at their respective required resolutions. For other methods, the input image resolution is uniformly set to 224x224.}
	\vspace{-3mm}
    \begin{tabular}{c c c c c c c c}
        \toprule
        Method & Abs Rel $\downarrow$ & Sq Rel $\downarrow$ & RMSE $\downarrow$ & RMSE(log) $\downarrow$ & $\delta < 1.25$ $\uparrow$ & $\delta < 1.25^2$ $\uparrow$ & $\delta < 1.25^3$ $\uparrow$\\
        \midrule
        FSM~\cite{guizilini2021surroundmonodepthmultiplecameras} & 0.334 & 2.845 & 7.786 & 0.406 & 0.508 & 0.761 & 0.894 \\
        R3D3~\cite{r3d3} & 0.253 & 4.759 & \textbf{7.150} & \textbf{0.347} & \textbf{0.729} & 0.848 & 0.903 \\
        SurroundDepth~\cite{wei2022surround} & 0.280 & 4.401 & 7.467 & 0.364 & 0.661 & 0.844 & \textbf{0.917} \\
        DUSt3R~\cite{dust3r_cvpr24} & 0.254 & 2.824 & 7.455 & 0.362 & 0.652 & 0.798 & 0.897 \\
        MonST3R~\cite{zhang2024monst3r} & 0.238 & 2.525 & 7.245 & 0.350 & 0.674 & 0.835 & 0.899 \\
        Spann3R~\cite{wang20243d} & \textbf{0.229} & 2.714 & 7.313 & 0.358 & 0.661 & 0.818 & 0.885 \\
        Driv3R & 0.234 & \textbf{2.279} & 7.298 & 0.353 & 0.697 & \textbf{0.850} & 0.905 \\
        \bottomrule
    \end{tabular}
    \label{tab:depth_metrics_result}
	\vspace{-1mm}
\end{table*}

\begin{table*}[t]
    \centering
    \caption{\textbf{Comparison of reconstruction results on nuScenes~\cite{nuscenes2019} dataset.} To comprehensively evaluate the model performance on reconstructing dynamic scenes, we sample 3508 input sequences from the NuScenes validation dataset according to the dynamic masks from the 4D flow predictor for evaluation. The resolution settings are the same as in Table~\ref{tab:depth_metrics_result}.}
	\vspace{-3mm}
    \begin{tabular}{c c c c c c c c c c c}
        \toprule
        \multirow{2}{*}{Methods} & \multicolumn{2}{c}{\textbf{Acc} $\downarrow$} & \multicolumn{2}{c}{\textbf{Comp} $\downarrow$} & \multicolumn{2}{c}{\textbf{NC} $\uparrow$} & \multicolumn{4}{c}{\textbf{Depth Metrics}} \\
        \cmidrule(lr){2-3} \cmidrule(lr){4-5} \cmidrule(lr){6-7} \cmidrule(lr){8-11}
        & Mean & Median & Mean & Median & Mean & Median & Abs Rel & Sq Rel & RMSE & $\delta < 1.25$ \\
        \midrule
        FSM~\cite{guizilini2021surroundmonodepthmultiplecameras} & 3.630 & 2.599 & 2.400 & 1.748 & 0.563 & 0.644 & 0.359 & 3.465 & 7.968 & 0.494 \\
        R3D3~\cite{r3d3} & 2.454 & 1.732 & 1.458 & 0.740 & 0.605 & 0.668 & 0.250 & 4.452 & 7.080 & 0.697 \\
        SurroundDepth~\cite{wei2022surround} & 2.873 & 1.964 & 1.620 & 0.875 & 0.613 & 0.670 & 0.288 & 4.560 & 7.677 & 0.655 \\
        DUSt3R~\cite{dust3r_cvpr24} & 2.840 & 2.026 & 1.377 & 0.682 & 0.627 & 0.708 & 0.325 & 3.780 & 7.682 & 0.604 \\
        MonST3R~\cite{zhang2024monst3r} & 1.635 & 1.247 & 1.469 & 0.714 & 0.640 & 0.712 & 0.240 & \textbf{2.025} & 6.784 & 0.689 \\
        Spann3R~\cite{wang20243d} & 2.372 & 1.726 & 1.334 & 0.699 & 0.636 & 0.716 & 0.293 & 3.202 & 7.397 & 0.636 \\
        Driv3R & \textbf{1.619} & \textbf{1.137} & \textbf{1.050} & \textbf{0.510} & \textbf{0.642} & \textbf{0.724} & \textbf{0.229} & 2.040 & \textbf{6.455} & \textbf{0.709} \\
        \bottomrule
    \end{tabular}
    \label{tab:main_result}
	\vspace{-5mm}
\end{table*}

\subsection{Results and Analysis} \label{sec:results}

\begin{figure*}[t]
    \centering
    \includegraphics[width=1\linewidth]{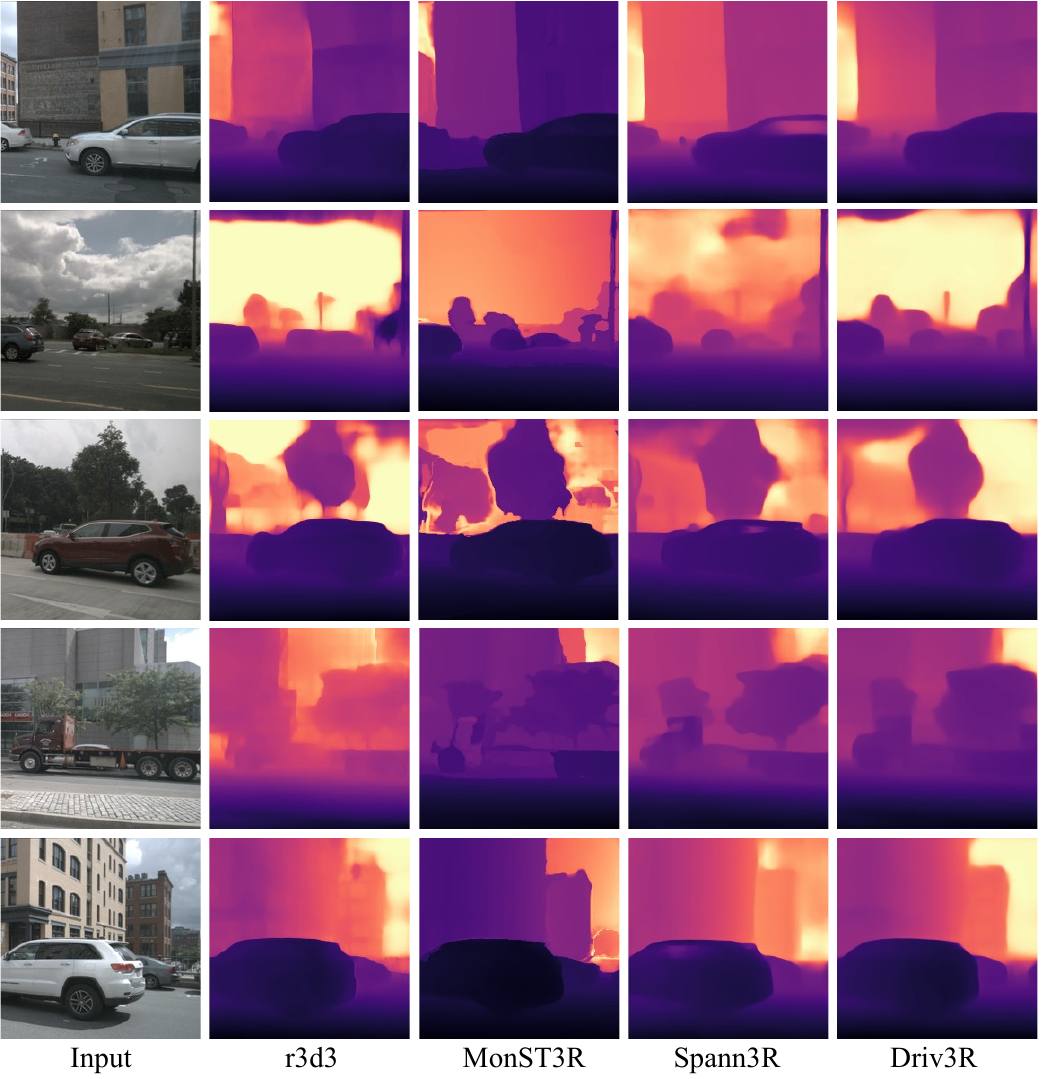}
	\vspace{-7mm}
    \caption{\textbf{Visualization Results of Depth Prediction on nuScenes~\cite{nuscenes2019} dataset.} In these cases, Driv3R leverages the R3D3~\cite{r3d3} model to mitigate the blur in Spann3R~\cite{wang20243d} caused by fast motion, resulting in more precise reconstruction of the dynamic scenes.}
	\vspace{-5mm}
    \label{fig:depth_vis}
\end{figure*}

\textbf{Depth Estimation.} Table~\ref{tab:depth_metrics_result} shows that our Driv3R achieves competitive results on nuScenes~\cite{nuscenes2019} compared to both multi-view depth estimation frameworks~\cite{guizilini2021surroundmonodepthmultiplecameras, r3d3, wei2022surround} and methods that directly regress per-frame pointmap~\cite{dust3r_cvpr24, zhang2024monst3r, wang20243d}. Notably, the visual results in Figure~\ref{fig:depth_vis} demonstrate that the spatial memory pool introduced in Spann3R~\cite{wang20243d} is not capable of fully capturing the temporal relationships within the input sequences, thus can cause significant blur in the reconstruction areas of fast-moving dynamic objects. Additionally, while MonST3R~\cite{zhang2024monst3r} can achieve more delicate depth estimation in some scenarios, its global alignment process is both computationally expensive (as discussed in Section~\ref{sec:results}) and highly sensitive to the accuracy of dynamic masks. In contrast, Driv3R leverages the strengths of Spann3R in static reconstruction and R3D3~\cite{r3d3} in depth estimation of dynamic objects, therefore is more efficient and robust for 4D dynamic reconstruction.

\begin{figure*}[t]
    \centering
    \includegraphics[width=1\linewidth]{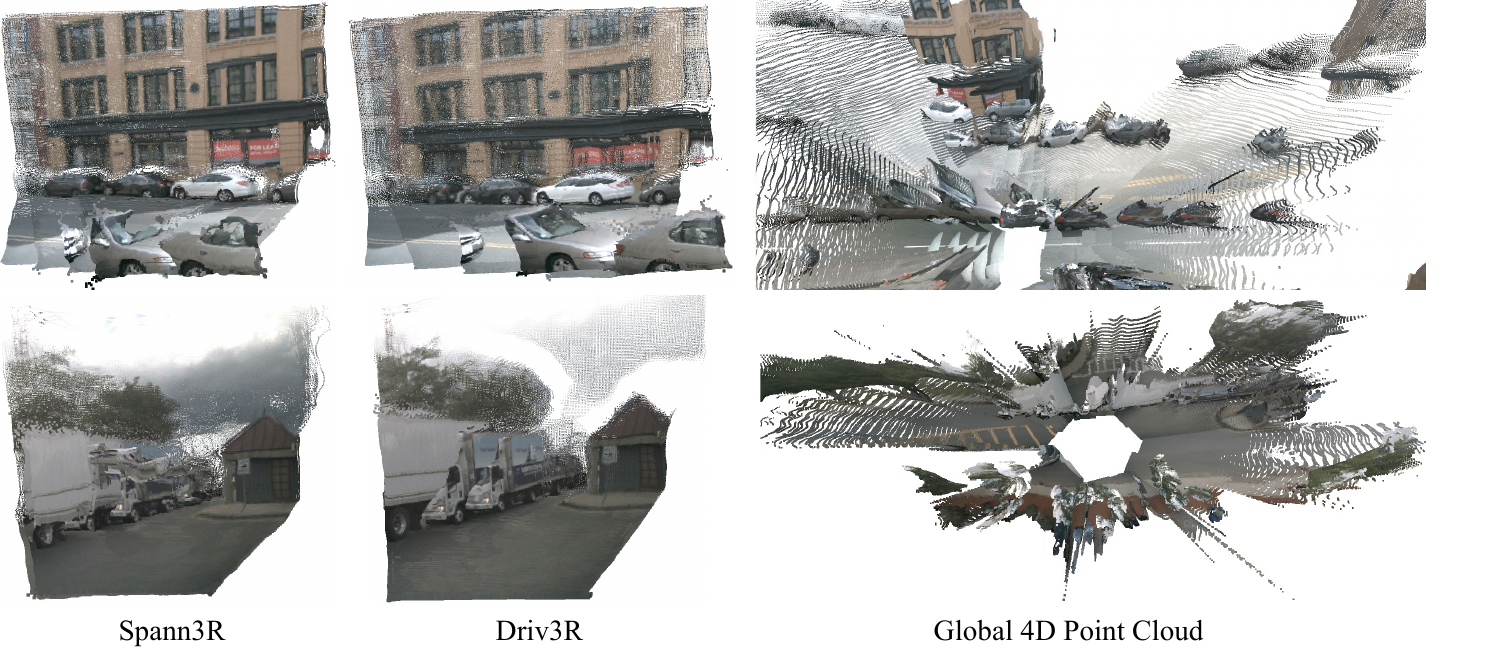}
	\vspace{-8mm}
    \caption{\textbf{Visualization Results of 4D Reconstruction on nuScenes~\cite{nuscenes2019}.} Driv3R leverages both strong 3D priors from DUSt3R~\cite{dust3r_cvpr24} and accurate dynamic predictions in R3D3~\cite{r3d3}, while maintaining both temporal and spatial consistency on 4D reconstruction.}
	\vspace{-6mm}
    \label{fig:pcd_vis}
\end{figure*}

\noindent\textbf{Dynamic Objects and Scene Reconstruction.} To highlight the ability of Driv3R to accurately reconstruct dynamic scenes, we sample 3,508 input sequences from the nuScenes~\cite{nuscenes2019} validation set, which contain a higher number of dynamic objects identified by the 4d flow predictor. As shown in Table~\ref{tab:main_result}, Driv3R outperforms all previous methods in both reconstruction and depth estimation on the NuScenes dynamic subset. Notably, our model even achieves slightly better reconstruction results on dynamic scenes compared to MonST3R~\cite{zhang2024monst3r}, which requires global alignment with flow optimization, while performing inference far more efficiently without the optimization. Visualization results further demonstrate that the point clouds of fast-moving objects predicted by Spann3R~\cite{wang20243d} often suffer from incompleteness, blurring, and inaccuracy. In comparison, our Driv3R reconstructs the dynamic regions of fast-moving objects more accurately due to the temporal interaction with the memory pool and the guidance of the dynamic masks from the 4D flow predictor. Furthermore, the 4D global point cloud generated by the optimization-free multi-view aligner shown in Figure~\ref{fig:pcd_vis} maintains strong 3D consistency as the encoded features from different viewpoints can share spatial information within the memory pool. Due to the point maps from Driv3R still containing floaters on the edges, we only retain the regions with higher confidence and align them to the global coordinate system via the optimization-free multi-view aligner for visualization.

\textbf{Efficiency Analysis.}
We investigate the inference efficiency of Driv3R in comparison to dominant methods that regress per-frame point maps or depth maps from input images. The input resolution is set to 224x224 for DUSt3R~\cite{dust3r_cvpr24}, MonST3R~\cite{zhang2024monst3r}, Spann3R~\cite{wang20243d} and Driv3R, while For R3D3~\cite{r3d3}, FSM~\cite{guizilini2021surroundmonodepthmultiplecameras}, and SurroundDepth~\cite{wei2022surround}, we evaluate the models at their respective required resolutions. All models are evaluated on a single A6000 GPU with a batch size of 1. To assess inference FPS, we use image sequences from a single sensor as input and also account for the time consumed by global alignment when evaluating inference time for DUSt3R and MonST3R. Additionally, we set the input sequence length to 5 and evaluate the total reconstruction time $t_{4d}$ and memory required to reconstruct the complete 4D scene from multi-view images. For DUSt3R, MonST3R, and Spann3R, we simply merge point maps from different sensors to assess the inference time for 4D reconstruction.

\begin{table}[t]
    \centering
    \caption{\textbf{Efficiency Analysis on nuScenes~\cite{nuscenes2019} dataset.} Time consumption of the global alignment process in DUSt3R~\cite{dust3r_cvpr24} and MonST3R~\cite{zhang2024monst3r} is also considered to be part of the inference time.}
	\vspace{-3mm}
    \begin{tabular}{c c c c}
        \toprule
        Method & FPS & Memory (GB) & $t_{4d}$ (s) \\
        \midrule
        DUSt3R~\cite{dust3r_cvpr24} & 0.40 & 5.84 & 178.31 \\
        MonST3R~\cite{zhang2024monst3r} & 0.19 & 5.98 & 312.44 \\
        Spann3R~\cite{wang20243d} & 4.63 & 5.82 & 12.98 \\
        FSM~\cite{guizilini2021surroundmonodepthmultiplecameras} & 2.45 & 4.60 & 29.39 \\
        R3D3~\cite{r3d3} & 2.79 & 2.96 & 21.51 \\
        SurroundDepth~\cite{wei2022surround} & 3.04 & 3.87 & 19.60 \\
        Driv3R & 4.55 & 7.84 & 13.18 \\
        \bottomrule
    \end{tabular}
    \label{tab:efficiency_result}
	\vspace{-4mm}
\end{table}

Table~\ref{tab:efficiency_result} shows that our Driv3R significantly surpasses DUSt3R~\cite{dust3r_cvpr24} and MonST3R~\cite{zhang2024monst3r}, which both require global alignment optimization, in terms of inference efficiency. Additionally, Driv3R also achieves better efficiency compared to methods that rely on monocular depth prediction~\cite{monodepth2} or multi-view fusion~\cite{r3d3, wei2022surround} for scene reconstruction. While our model consumes more inference memory due to the additional memory storage, it achieves efficiency comparable to Spann3R~\cite{wang20243d} where we simply merge the pointmaps from different sensors to formulate the 4D representations. However, Driv3R goes further by accurately identifying and modeling dynamic objects within the 4D flow predictor, and ensuring both temporal and spatial consistency through interactions with the temporal-spatial memory pool, thus achieving superior performance for 4D reconstruction in dynamic scenes.

\label{sec:4.4}

\textbf{Ablation of modules.}
To further investigate the designs of Driv3R, we conduct ablations on the nuScenes~\cite{nuscenes2019} dynamic subset. We begin with a vanilla model without the temporal-spatial memory pool or the 4D flow predictor. Next, we introduce temporal feature interactions within the memory pool. Furthermore, the memory pool is extended, and our model performs both temporal interactions across input sequences and spatial interactions between different viewpoints. Finally, we incorporate the 4D flow predictor to generate masks for dynamic objects within the scene. All the models are trained based on the Spann3R~\cite{wang20243d} pretrained weight for 50 epochs and take the point predictions from the R3D3~\cite{r3d3} model as supervision. Note that we perform the global optimization on the vanilla model to align all point maps to a consistent coordinate system.

\begin{table}[t]
    \centering
    \caption{\textbf{Ablations on the nuScenes~\cite{nuscenes2019} dataset.} We report the standard metrics for depth estimation, including the average Abs Rel, Sq Rel, and RMSE, respectively.}
	\vspace{-3mm}
    \begin{tabular}{c c c c}
        \toprule
        Model & Abs Rel & Sq Rel & RMSE \\
        \midrule
        vanilla & 0.325 & 3.780 & 7.682 \\
        + temporal interactions & 0.293 & 3.202 & 7.397 \\
        + spatial interactions & 0.268 & 2.875 & 7.034 \\
        + 4d flow predictor & \textbf{0.229} & \textbf{2.040} & \textbf{6.455} \\
        \bottomrule
    \end{tabular}
	\vspace{-6mm}
    \label{tab:ablation}
\end{table}

As shown in Table~\ref{tab:ablation}, while temporal interactions across timestamps improve the reconstruction quality compared to using the direct DUSt3R encoder-decoder architecture, introducing spatial interactions across viewpoints (as shown in Figure~\ref{fig:spatial_memory_pool}) results in a further boost to model performance. Additionally, dynamic masks generated by the 4D flow predictor mitigate the impact of inaccurate predictions from the R3D3~\cite{r3d3} model. This allows Driv3R to leverage both the strong 3D priors of DUSt3R~\cite{dust3r_cvpr24} for static scenes and representations of moving objects predicted from R3D3 multi-view estimation, resulting in significantly improved performance in dynamic 4D reconstruction.

\section{Conclusion}

We have presented Driv3R to learn dense 4D reconstruction on dynamic scenes for autonomous driving. Our key innovation is a memory pool that reasons both temporal relationships across sequences and spatial contexts across viewpoints. We also use a 4D Flow Predictor to identify moving objects, guiding the network to focus on dynamic regions. With an optimization-free multi-view aligner, Driv3R generates consistent 4D point maps in the global coordinate system. On the large-scale NuScenes~\cite{nuscenes2019} dataset, Driv3R outperforms existing methods in depth estimation and scene reconstruction, achieving $15 \times$ higher inference speed than methods relying on global alignment.

\vspace{0.05cm}

\noindent\textbf{Limitations.} Although Driv3R efficiently reconstructs large-scale dynamic scenes, the input length is constrained by memory. During training, it requires about 10 GB of memory for a 5-frame sequence from 6 multi-view cameras, mainly due to memory pool storage. Furthermore, using sparse LiDAR points for dynamic objects as supervision does not yield optimal results, and point predictions from the R3D3~\cite{r3d3} pre-trained model can be inaccurate in some cases. Future work can focus on improving memory storage and adapting to the fully self-supervised training.

\appendix
\section{Additional Implementation Details}

\subsection{Training Configuration} \label{app:training}

The comprehensive configuration for the two-stage training of Driv3R is shown in Table~\ref{tab:experimental_settings}. Based on Spann3R~\cite{wang20243d} pretrained weight, we first train Driv3R for 30 epochs using only temporal sequences as input. Next, we fine-tune the model for 20 more epochs with a smaller learning rate. In this stage, temporal sequences from multiple sensors are adopted as input, allowing the memory pool to further refine its reasoning of spatial contexts. For all the baseline models that have not been trained on nuScenes~\cite{nuscenes2019} dataset, we fine-tune their pretrained model with the same training configuration for a fair comparison.

\vspace{-2mm}
\begin{table}[h]
    \centering
    \caption{Detailed configuration of the two-stage training process for Driv3R on the large-scale nuScenes~\cite{nuscenes2019} dataset.}
	\vspace{-3mm}
    \begin{tabular}{c|c|c}
    \toprule
    Config & Temporal & Spatial \\ \midrule
    optimizer & AdamW & AdamW \\
    scheduler & OneCycleLR & OneCycleLR \\
    learning rate & $3 \times 10^{-5}$ & $2 \times 10^{-5}$ \\
    weight decay  & 0.05 & 0.05 \\
    batch size & 4 & 4 \\
    epoch & 30 & 20 \\
    \bottomrule
    \end{tabular}
	\vspace{-5mm}
    \label{tab:experimental_settings}
\end{table}

\subsection{Training Loss}

As illustrated before, the training loss is a combination of confidence loss and scale loss. Following DUSt3R~\cite{dust3r_cvpr24}, the confidence loss is defined as:
\begin{equation}
\label{eq:confidence_loss}
    \mathcal{L}_{conf} = \sum_{c} \sum_{t} \sum_{i \in M_{t, c}} C_{t, c}^{i} \ \mathcal{L}_{reg}(i) - \alpha \ log \ C_{t, c}^{i},
\end{equation}
\begin{equation}
\label{eq:scale_loss}
    \mathcal{L}_{scale} = max(0, X(M_{t, c}) - X_{r3d3}(M_{t, c})),
\end{equation}
where $M_{t, c}$ represents the dynamic mask of timestamp $t$ and sensor $c$ generated from the 4D Flow Predictor, while $\alpha$ controls the overall range of confidence score and is set to 0.5 during the training. Additionally, the scale loss is formulated as Equation~\ref{eq:scale_loss}, which guides our Driv3R model to predict points of dynamic objects within a narrower range compared to the ``ground truth" predictions from the R3D3~\cite{r3d3} model, thereby preventing the predictions of fast-moving objects from becoming unstable.

\subsection{Model Architecture}

In this section, we further elaborate on the implementation details of the temporal-spatial memory pool in Driv3R. Different from Spann3R~\cite{wang20243d}, we maintain a separate memory pool for each sensor where key-value pairs are stored in chronological order based on their timestamps. During the first stage of training, the features of the input sequences are updated using only the memory pool corresponding to their respective sensor. However, during the spatial training stage, we first pre-define the frames from the previous $k$ timestamps (set to 4 in our experiment) that overlap with the field of view of the current frame as the most closely related frames. Then, we apply cross-attention to update the features of the current frame using only these closely related frames to address the limitations of training memory.

Furthermore, we maintain working memory and long-term memory for each pool inspired by Spann3R~\cite{wang20243d}. The working memory consists of the most recent 5 frames during the first training stage, and the most similar 5 frames during the second spatial training stage. For long-term memory, we apply memory pruning based on the accumulated attention weights, preventing the memory from exceeding its limit during large-scale training and inference.

\section{Additional Results and Visualization}

\begin{figure*}[t]
    \centering
    \includegraphics[width=\linewidth]{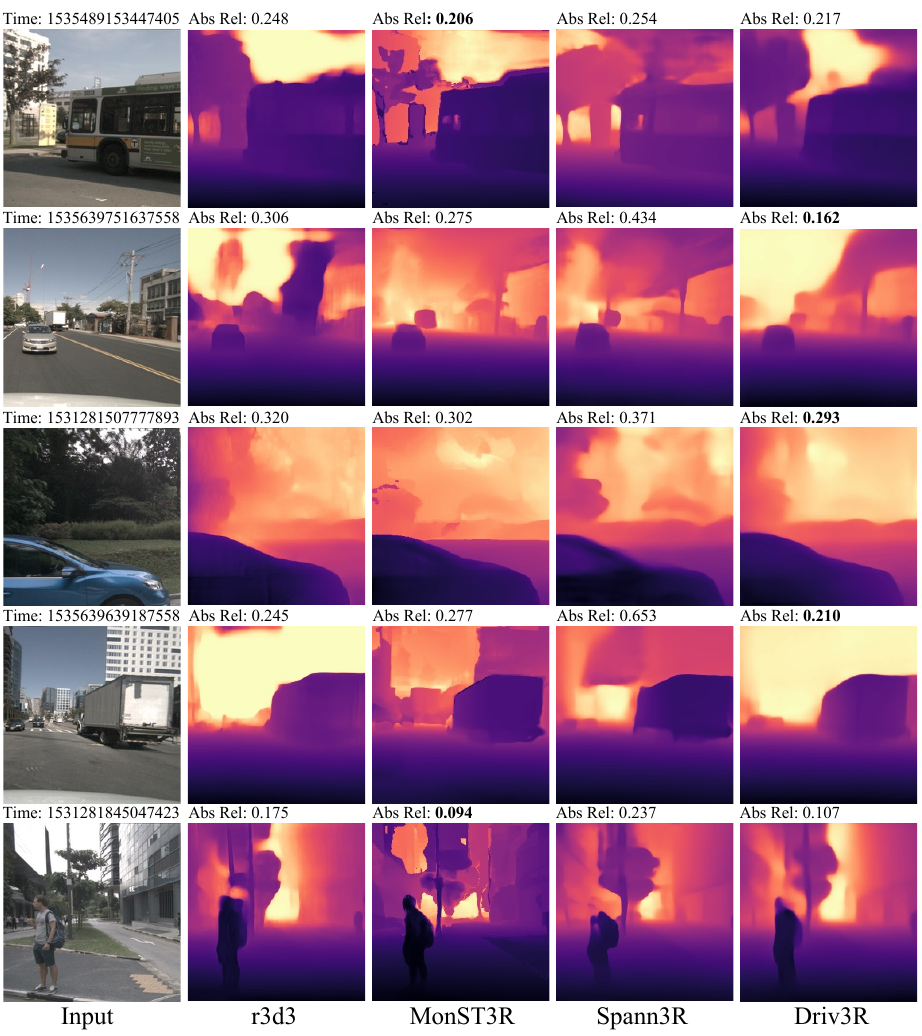}
	\vspace{-7mm}
    \caption{\textbf{Addtional Visualization Results of Depth Prediction on nuScenes~\cite{nuscenes2019} dataset.} While MonST3R~\cite{zhang2024monst3r} demonstrates more refined depth prediction in certain scenarios, Driv3R delivers comparable performance with a 15x faster inference speed, outperforming methods which rely on global alignment optimization to reconstruct large-scale dynamic scenes.}
    \label{fig:add_depth_vis}
	\vspace{-4mm}
\end{figure*}

\begin{figure*}[t]
    \centering
    \includegraphics[width=0.95\linewidth]{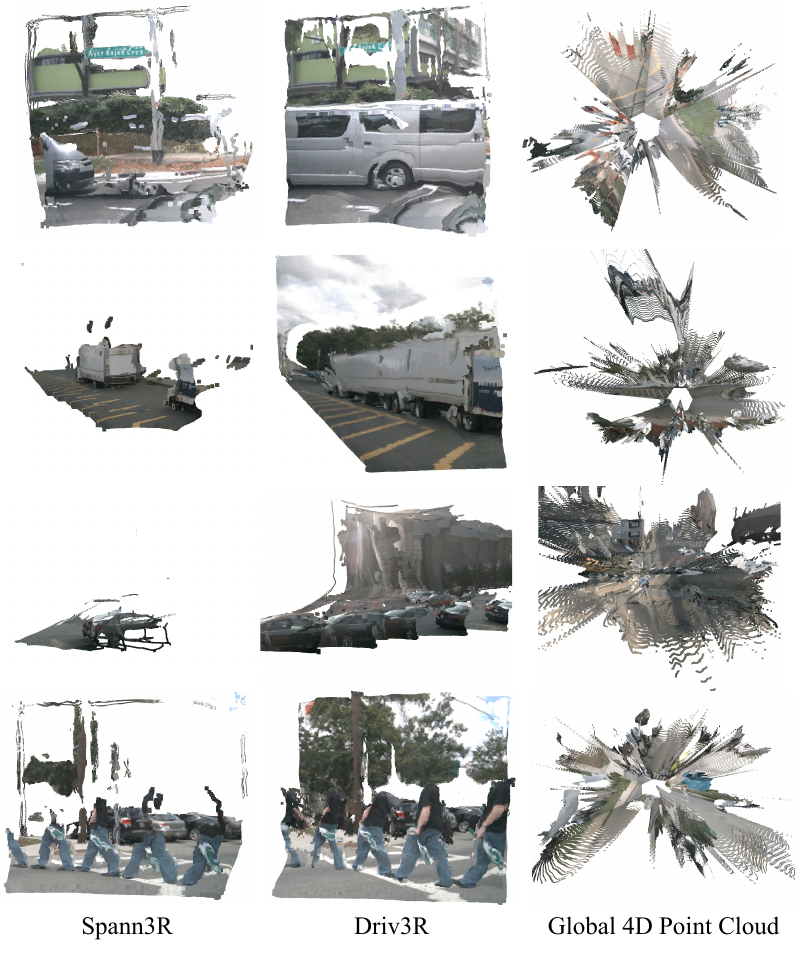}
	\vspace{-3mm}
    \caption{\textbf{Additional Visualization Results of 4D reconstruction on nuScenes~\cite{nuscenes2019} dataset.} We only preserve points with confidence values larger than a predefined threshold to ensure a reliable and fair comparison.}
    \label{fig:add_pcd}
	\vspace{-5mm}
\end{figure*}

We present more visualization results of the depth maps and 4D global point clouds with corresponding quantitative results for Driv3R, Spann3R~\cite{wang20243d}, MonST3R~\cite{zhang2024monst3r}, and R3D3~\cite{r3d3} on the large-scale nuScenes~\cite{nuscenes2019} dataset. As illustrated in Figure~\ref{fig:add_depth_vis}, our Driv3R effectively integrates the strengths of R3D3~\cite{r3d3} for dynamic object prediction and Spann3R~\cite{wang20243d} for static background modeling, achieving comparable depth prediction while offering a 15x faster inference speed compared to MonST3R~\cite{zhang2024monst3r}. Notably, our model outperforms MonST3R in several scenarios due to its robustness in handling the inaccuracy of dynamic masks. Furthermore, we provide more visualizations of local and global 4D reconstruction in Figure~\ref{fig:add_pcd}, where Driv3R achieves better representations of fast-moving objects and ensures both temporal and spatial consistency.

\vspace{-0.1cm}

\section{Additional Discussion and Future works}

Although Driv3R outperforms previous methods in the 4D reconstruction of large-scale dynamic scenes, the point predictions still exhibit floating artifacts on the edges, and there are several failure cases involving both fast-moving objects and the background. These limitations primarily arise from the accuracy constraints of the point predictions in the R3D3~\cite{r3d3} model, as well as the lack of high-quality and continuous data sequences containing a large number of dynamic and fast-moving objects. Due to the lack of ground-truth dense point clouds for autonomous driving and inaccurate depth prediction of existing methods, our future research will focus on adapting our model to an efficient self-supervised approach, such as utilizing reprojection loss and geometry consistency loss. Furthermore, to address the issue of data scarcity, a more efficient data sampling strategy should be adopted instead of splitting sequence data based on temporal order. Moreover, future work could involve joint training strategies on multiple large-scale datasets, even those outside the context of autonomous driving, to further improve the model performance and robustness.

\end{document}